# MOBILE ROBOT LOCALIZATION USING FUZZY NEURAL NETWORK BASED EXTENDED KALMAN FILTER

**Thi Thanh Van Nguyen, Manh Duong Phung, Thuan Hoang Tran, Quang Vinh Tran**

*University of Engineering and Technology, Vietnam National University, Hanoi*
*144 Xuan Thuy, Cau Giay, Hanoi, Vietnam*
Corresponding author*: vanntt@vnu.edu.vn*

## ABSTRACT

This paper proposes a novel approach to improve the performance of the extended Kalman filter (EKF) for the problem of mobile robot localization. A fuzzy logic system is employed to continuously adjust the noise covariance matrices of the filter. A neural network is implemented to regulate the membership functions of the antecedent and consequent parts of the fuzzy rules. The aim is to gain the accuracy and avoid the divergence of the EKF when the noise covariance matrices are fixed or wrongly determined. Simulations and experiments have been conducted. The results show that the proposed filter is better than the EKF in localizing the mobile robot.

## I. INTRODUCTION

Localization, that is the determination of the robot's position and orientation from sensor data, is a fundamental problem in mobile robotics. In order to complete given tasks, the robot need to know its own pose at each sampling time to make the path planning and motion control. The difficulty is that there always exist two kinds of error in the robot system: the systematic and non systematic errors [1]. The systematic error is caused by the imperfectness of robot mechanics such as the limitation of encoder resolution, the ansymmetry of robot chassis, and the inequality and misalignment of driven wheels. The non systematic error, on the other hand, is often random and unknown. It is caused by uncertainty elements such as the slippage of wheels or the unevenness of floor during the robot operation. Filtering these errors to extract the actual pose of the robot is the final goal of localization algorithms. Various approaches have been proposed with their strengths and weaknesses.

In [2], one linear and two rotary encoders are used to measure the relative distance and bearing between two wheels. The result is then injected to the conventional dead reckoning method to improve the accuracy. Another approach is the Monte Carlo method that is able to localize the mobile robot without knowledge of its starting location. This method is more accurate, faster, and less memory intensive than grid based methods [3]. A survey of Bayesian filter applied to real world estimation was done by authors of university of Washington [4]. This research shows that the Bayesian filter technique is a powerful statistical tool to perform the multi-sensor fusion, estimate the system state and manage the measurement uncertainties. Its drawback is the expensive computation.

A less computation but effective method is the extended Kalman filter (EKF) [5]. It estimates the robot position under the scenaio that both sensor and system data are subjected to zero-mean Gaussian white noises [5, 6]. In this method, the choice of noise covariance matrices greatly affects the estimate accuracy. Due to random essence of errors, these matrices change according to the time of operation and therefore are difficult to determine. In practice, these matrices are often assumed to be fixed and chosen through off-line processes. However, this simplification may cause the EKF to diverge in some cases.

In this paper, a fuzzy system is implemented to online adjust the noise covariance matrices at each Kalman step. The membership functions of the fuzzy system are regulated by a neural network. The incorporation of fuzzy logic and neural network into the EKF creates a new optimal filter called the fuzzy neural network based EKF (FNN-EKF). This filter enhances the accuracy and convergence of the EKF for the problem of localization in unknown indoor environment. This combined with previous results in path planning and obstacle avoidance of the author's group constitute a quite complete set of mechanisms for the autonomous operation of a mobile robot [7-9].

## II. MOBILE ROBOT LOCALIZATION USING EXTENDED KALMAN FILTER

As mentioned in previous section, the EKF is considered as one of the most effective method for mobile robot localization. This section first presents the kinematic model of the robot. The implementation of the EKF is then described and discussed.

### *A. Robot model*

The mobile robot considered in this research is the two-wheeled, differential-drive mobile robot with non-slipping and pure rolling. Figure 1 shows the coordinate system of the robot, where ($X_G$, $Y_G$) is the global coordinate system and ($X_R$, $Y_R$) is the local coordinate system relative to the robot chassis. With this kind of mobile robot, the dead reckoning is often used to determine the relative position of the robot in the work space. It estimates the robot's position and orientation based on encoder data.

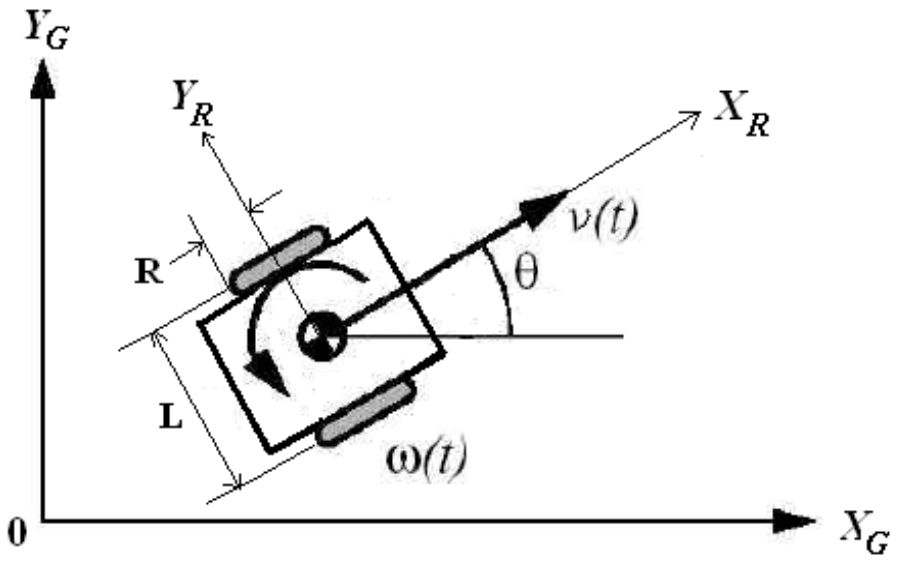

Figure 1: Pose and parameters of mobile robot

From figure 1, the conversion factor that translates encoder pulses into linear wheel displacement is given by:

$$C_m = \frac{\pi R}{n C_e} \qquad (1)$$

where $R$ is the wheel diameter, $n$ is the gear ratio of the reduction gear between the motor and the drive wheel, and $C_e$ is the encoder resolution (in pulses per revolution).

Let $N_{L,i}$ and $N_{R,i}$ be the number of pulses counted by encoders of the left and right wheels at time $i$, respectively. The displacement of each wheel is then given by:

$$\Delta S_{L,i} = C_m N_{L,i} \qquad \Delta S_{R,i} = C_m N_{N,i} \qquad (2)$$

These can be translated to the linear incremental displacement of the robot's center $\Delta S$ and orientation angle $\Delta\theta$ as:

$$\Delta S_i = \frac{\Delta S_{L,i} + \Delta S_{R,i}}{2} \qquad \Delta\theta_i = \frac{\Delta S_{R,i} - \Delta S_{L,i}}{L} \qquad (3)$$

where $L$ is the distance between two wheels. The coordinates of the robot in the global coordinate frame then can be update by:

$$\begin{aligned} x_{i-1} &= x_{i-1} + \Delta S_{i-1} * \cos(\theta_{i-1} + \Delta\theta_{i-1}) \\ y_i &= y_{i-1} + \Delta S_i * \sin(\theta_{i-1} + \Delta\theta_{i-1}) \\ \theta_i &= \theta_{i-1} + \Delta\theta_i \end{aligned} \qquad (4)$$

From equation (4), it is able to determine the robot's position and orientation if we know the number of encoder pulses in each sampling period. In another word, equation (4) is the localization equation of the dead reckonging method. Nevertheless, equation (4) does not include errors appeared in the system. As analyzed in section I, errors are unavoidable and may downgrade the system per-formance if appropriate compensation is not investigated. For this reason, the robot model is rewritten in state-space representation with the appearance of disturbances as follows.

$$\mathbf{x}_i = f(\mathbf{x}_{i-1}, \mathbf{u}_{i-1}, \mathbf{w}_{i-1}) \tag{5}$$

$$\mathbf{z}_i = h(\mathbf{x}_i, \mathbf{v}_i) \tag{6}$$

where $\mathbf{x} = \begin{bmatrix} x_i & y_i & \theta_i \end{bmatrix}^T$ is the state vector described the instantaneous position and orientation of the robot, $\mathbf{u} = [\Delta S_{L,i} \ \ \Delta S_{R,i}]^T$ is the input, $\mathbf{z}_i$ is the measurments, $f$ and $h$ are the system functions, $\mathbf{w}_i$ and $\mathbf{v}_i$ are random variables described the process and measurement noises. These noises are assumed to be zero-mean, independent, white, and Gaussian with the process noise covariance matrix $\mathbf{Q}$ and the measurement noise covariance matrix $\mathbf{R}$ : $\mathbf{w}_i \sim \mathbf{N}(0, \mathbf{Q}_i)$, $\mathbf{v}_i \sim \mathbf{N}(0, \mathbf{R}_i)$. The state-space presentation (5) - (6) is basis for the implementation of the EKF.

*B. Implementation of the EKF for mobile robot localization*

As $\mathbf{x}$ is the state vector described the robot postion and orientation, the problem of localization becomes the problem of state estimation. An optimal solution is the Kalman filter [12]. Its implementation is performed through two steps: prediction and correction, as follows:

1. Prediction step with time update equations:

$$\hat{\mathbf{x}}_i^- = f(\hat{\mathbf{x}}_{i-1}, \mathbf{u}_{i-1}, 0) \tag{7}$$

$$\mathbf{P}_i^- = \mathbf{A}_i \mathbf{P}_{i-1} \mathbf{A}_i^T + \mathbf{W}_i \mathbf{Q}_{i-1} \mathbf{W}_i^T \tag{8}$$

where $\hat{\mathbf{x}}_i^-$ is the prior state estimate at step i given knowledge of the process prior to step $i$-1, $\mathbf{P}_i^-$ is the covariance matrix of the state prediction error, $\mathbf{P}_i$ is the covariance matrix of the corresponding estimation error, $\mathbf{A}$ is the Jacobian matrix of partial derivates $f$ to $\mathbf{x}$, $\mathbf{W}$ the Jacobian matrix of partial derivates $f$ to $\mathbf{w}$, $\mathbf{Q}$ the input noise covariance matrix.

2. Correction step with measurement update equations:

$$\mathbf{K}_i = \mathbf{P}_i^- \mathbf{H}_i^T (\mathbf{H}_i \mathbf{P}_i^- \mathbf{H}_i^T + \mathbf{V}_i \mathbf{R}_i \mathbf{V}_i^T)^{-1} \tag{9}$$

$$\hat{\mathbf{x}}_i = \hat{\mathbf{x}}_i^- + \mathbf{K}_i (\mathbf{z}_i - h(\hat{\mathbf{x}}_i^-, 0)) \tag{10}$$

$$\mathbf{P}_i = (\mathbf{I} - \mathbf{K}_i \mathbf{H}_i) \mathbf{P}_i^- \tag{11}$$

where $\hat{\mathbf{x}}_i$ is the posterior state estimate at step i given measurement $z_i$, $\mathbf{K}_i$ is the Kalman gain, $\mathbf{R}$ is the covariance matrix of measurement noise, $\mathbf{H}$ is the Jacobian matrix of partial derivates $h$ to $\mathbf{x}$, $\mathbf{V}$ is the Jacobian matrix of partial derivates $h$ to $\mathbf{v}$.

Equations (7) - (11) show that the noise covariance matrices $\mathbf{Q}$ and $\mathbf{R}$ need be accurately determined in order to ensure the efficient operation of the EFK. In addition, these matrices should be changed at each step of the EKF. The dynamic determination of $\mathbf{Q}$ and $\mathbf{R}$ however is often not easy to perform, especially during the operating session of the mobile robot. In practice, $\mathbf{Q}$ and $\mathbf{R}$ are often assumed to be fixed. Their values are determined before the execution of the EKF. This approach has several drawbacks. Firstly, fixed matrices $\mathbf{Q}$ and $\mathbf{R}$ do not reflect the variation of noises. Secondly, the covariance error $\mathbf{P}$ and the Kalman gain $\mathbf{K}$ will quickly reach the steady state. This is equivalent to the convergence of the EKF to certain degree accuracy. Finally, the fixation of $\mathbf{Q}$ and $\mathbf{R}$ may break the stability operation of the EKF and cause the system to be halted in some cases. To overcome these, fuzzy logic [13, 14] can be employed to enable an online adjustment (not the determination) of $\mathbf{Q}$ and $\mathbf{R}$ . In addition, the effeciciency of fuzzy rules can be futher improved by using neural network to regulate its membership functions.

## III. IMPROVEMENT EKF BY FUZZY NEURAL NETWORK SYSTEM

This section first presents the basis idea to adjust the covariance matrices **Q** and **R**. Details of implementation this idea using fuzzy logic and neural network are then described.

### A. *Concept of noise covariance matrices adjustment*

From equation (10), let $\mathbf{r}_i = \mathbf{z}_i - h(\hat{\mathbf{x}}_i^-, 0)$ be the residual between the actual and the predicted measurements. This residual, gained by **K**, is the correction factor to form the posterior estimate $\hat{\mathbf{x}}_i$ from the prior estimate $\hat{\mathbf{x}}_i^-$. It also reflects the accuracy of the estimation value. A small value of $\mathbf{r}_i$ implies good estimation as the predicted measurement is closed to the actual measurement. As mentioned in section II.B, if the covariance matrices $\mathbf{Q}$ and $\mathbf{R}$ are fixed, the Kalman gain $\mathbf{K}$ will quickly stabilize and remain constant. It means that the EKF will converge to certain accuracy degree. In order to improve this, we can adjust **Q** and **R** so that $\mathbf{r}_i$ is reduced.

Assume that the system operates with a predetermined and fixed $\mathbf{Q}$, the adjusting process of $\mathbf{R}$ to reduce $\mathbf{r}_i$ is performed as follows:

- Determining the residual $\mathbf{r}_i$ and computing its present covariance:

$$\mathbf{S}_i = \mathbf{H}_i \mathbf{P}_i^- \mathbf{H}_i^T + \mathbf{R}_i \tag{12}$$

- The average covariance of $\mathbf{r}_i$ through a number of steps from the past to present is given by:

$$\hat{\mathbf{C}}_i = \frac{1}{N} \sum_{j=j_0}^{i} \mathbf{r}_j \mathbf{r}_j^T \tag{13}$$

  where $j_0 = i$-N+1 and N is the number of past periods.

- The difference between the present and average covariances of $\mathbf{r}_i$ is given by:

$$\mathbf{D}_i = \mathbf{S}_i - \hat{\mathbf{C}}_i \tag{14}$$

  This difference is used as the reference to adjust $\mathbf{R}_i$.

- Let $\Delta R_i$ be an adjustment factor. $\Delta R_i$ is changed through each step as follows:

  1. *If* $\mathbf{D}_i \cong 0$ *then* maintain $\Delta R_i$
  2. *If* $\mathbf{D}_i > 0$ *then* decrease $\Delta R_i$
  3. *If* $\mathbf{D}_i < 0$ *then* increase $\Delta R_i$

- **R** is then adjusted as:

$$\mathbf{R}_i = \mathbf{R}_i + \Delta R_i \tag{15}$$

The adjustment of covariance matrix $\mathbf{Q}$ is similar to steps for $\mathbf{R}$ except that the present covariance matrix of $\mathbf{r}_i$ is replaced by:

$$\mathbf{S}_i = \mathbf{H}_i (\mathbf{A}_i \mathbf{P}_i \mathbf{A}_i^T + \mathbf{Q}_i) \mathbf{H}_i^T + \mathbf{R}_i \tag{16}$$

### B. *Fuzzy system implementation*

The fuzzy system to execute equation (15) is implemented through three steps as follows.

*Step 1:* Define input/output language variables

The input/output language variables of the fuzzy system are chosen as follows:

$\mathbf{D}_{i\,(j,\,j)}$: Negative (N), Zero (Z), Positive (P)

$\Delta R_i$ : Decrease (D), Maintain (M), Increase (I)

*Step 2:* Define membership functions

We define four sigmoid-shape membership functions $P(a_1,b_1)$, $N(a_2,b_2)$, $D(a_3,b_3)$ and $I(a_4,b_4)$ and two gauss-shape membership functions $M(c_1,\sigma_1)$ and $Z(c_2,\sigma_2)$ as follows:

$$sigmoid(x) = \frac{1}{1+e^{-a_i(x-b_i)}} \qquad gauss(x) = e^{\frac{-(x-c_i)^2}{2\sigma_i^2}} \tag{17}$$

where parameters $a_i$, $b_i$, $c_i$, and $\sigma_i$ are distinct values for each membership function.

*Step 3:* Design fuzzy rules

Rules for the fuzzy system are defined so that the difference between the present and average co-variances of the residual is reduced as follows:

1. If $\mathbf{D}_i$ is Z then $\Delta R_i$ is M
2. If $\mathbf{D}_i$ is P then $\Delta R_i$ is D
3. If $\mathbf{D}_i$ is N then $\Delta R_i$ is I

*Step 4:* Defuzzification

The defuzzification is accomplished by centroid method.

$$\phi = \frac{\sum x_i \mu(x_i)}{\sum \mu(x_i)} \tag{18}$$

where $x_i$ is the I'th domain value and $\mu(x_i)$ is the truth membership value for that domain point.

Among steps, the definition of membership functions requires manual customization. It is therefore time consuming and often do not adapt to the change of system parameters. In order to overcome this limitation, a neural network is implemented to regulate membership functions.

### C. The fuzzy neural network system

The objective of the neural network is to tune parameters $a_i$, $b_i$, $c_i$, and $\sigma_i$ so that the membership functions are fitter with the system. We design a neural network with five layers as shown in the figure 2. Details of each layer are as follows:

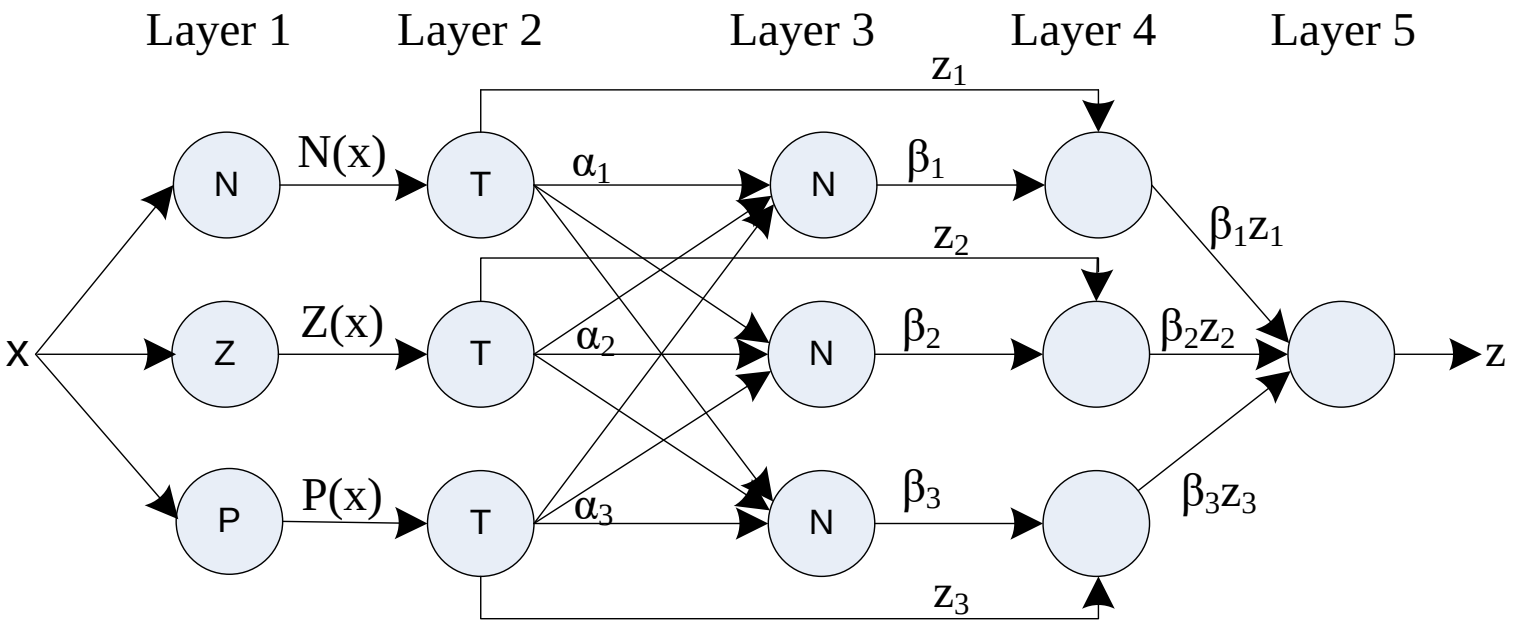

Figure 2: The fuzzy neural network

Layer 1: The output of the node is the degree of the variable with respect to fuzzy set N, Z and P.

Layer 2: The firing level of each rule is computed by

$$\alpha_1 = \mathrm{N}(x), \quad \alpha_2 = \mathrm{Z}(x), \quad \alpha_3 = \mathrm{P}(x) \tag{19}$$

Layer 3: The normalization of the firing levels is indicated. The normalized firing level of the corresponding rule is

$$\beta_1 = \frac{\alpha_1}{\alpha_1+\alpha_2+\alpha_3}, \beta_2 = \frac{\alpha_2}{\alpha_1+\alpha_2+\alpha_3}, \beta_3 = \frac{\alpha_3}{\alpha_1+\alpha_2+\alpha_3} \tag{20}$$

Layer 4: The product of the normalized firing level and the individual rule output of the corresponding rule is the output of each neuron. They are determined respectively as below

$$\beta_1 z_1 = \beta_1 \mathrm{I}^{-1}(\alpha_1), \beta_2 z_2 = \beta_2 \mathrm{M}^{-1}(\alpha_2), \beta_3 z_3 = \beta_3 \mathrm{D}^{-1}(\alpha_3) \tag{21}$$

Layer 5: The overall output system is the sum of all incoming signals

$$z = \beta_1 z_1 + \beta_2 z_2 + \beta_3 z_3 \tag{22}$$

With given the crisp training set $\{(D_1,\Delta R_1),(D_2,\Delta R_2)...(D_K,\Delta R_K)\}$, we define the measure of error for the k-th training pattern as

$$E_k = \frac{1}{2}(z_k - \Delta R_k)^2, \qquad k = 1...K \tag{23}$$

where $z_k$ is the computed output form of the fuzzy system corresponding to the input pattern $\mathbf{D}_k$ and the desired output $\Delta R_k$. The hybrid neural network learns the shape parameters $a_i$, $b_i$, $c_i$, and $\sigma_i$ of each membership function by using the steepest descent method. They are described as:

$$\begin{aligned} a_i(t+1) &= a_i(t) - \eta\frac{\partial E_k}{\partial a_i} & \qquad b_i(t+1) &= b_i(t) - \eta\frac{\partial E_k}{\partial b_i} \\ c_i(t+1) &= c_i(t) - \eta\frac{\partial E_k}{\partial c_i} & \qquad \sigma_i(t+1) &= \sigma_i(t) - \eta\frac{\partial E_k}{\partial \sigma_i} \end{aligned} \tag{24}$$

where $\eta>0$ is the learning constant and $t$ is the number of the adjustments.
Parameters extracted from equation (21) are used to construct new membership functions or learned membership functions for the fuzzy inference system.

## IV. SIMULATIONS

### A. *Simulation setup*

Simulations are conducted to evaluate the effeciency of the proposed algorithm. In each evaluation, a Monte Carlo simulation with 100 times is executed to compare the performance of the EKF and the FNN-EKF. Parameters for simulations are extracted from a real mobile robot built by the author's research group at the university's laboratory [15].

The mobile robot has following parameters: the wheel's diameter R is 0.05m, the distance between the wheels L is 0.6m, the gear ratio of the reduction gear between the motor and the drive wheel n equals to 1, and the encoder resolution $C_e$ is 500. The input noise covariance matrix **Q** is defined as propotional to the deviations of the displacement, $\sigma_{\Delta S}$, and the orientation, $\sigma_{\Delta\theta}$, of the robot. The covariance of measurement noise is chosen based on the value of the real robot system described in next section. (25) shows the values of **Q** and **R**.

$$\mathbf{Q} = \begin{bmatrix} \sigma_{\Delta S}^2 & 0 \\ 0 & \sigma_{\Delta\theta}^2 \end{bmatrix} \qquad \mathbf{R} = \begin{bmatrix} 0.01 & 0 & 0 \\ 0 & 0.01 & 0 \\ 0 & 0 & 0.018 \end{bmatrix} \tag{25}$$

Initial membership functions $P(a_1,b_1)$, $N(a_2,b_2)$, $D(a_3,b_3)$, $I(a_4,b_4)$, $M(c_1,\sigma_1)$ and $Z(c_2,\sigma_2)$ are chosen with parameters as follows: $a_1$ = -0.16, $b_1$ = -17, $a_2$ = 0.1, $b_2$ =13, $a_3$ = 0, $b_3$ = -100, $a_4$ = 0.1, $b_4$ = 23, $\sigma_1$ = 0, $c_1$ = 0.03, $\sigma_2$ = 0.03 and $c_2$ = 0.03. After the learning process of neural network, these parameters are as follows: $a_1$ = -0.1073, $b_1$ = -17.0995, $a_2$ = 0.1515, $b_2$ =13.8108, $a_3$ = -0.06552, $b_3$ = -143, $a_4$ = 0.1030, $b_4$ = 24.2000, $\sigma_1$ = 0.1779, $c_1$ = 0.0571, $\sigma_2$ = 0.0620 and $c_2$ = 0.0467. Figure 3 and figure 4 show the initial and learned shapes of the membership functions D, M and I of $\Delta R_i$, respectively.

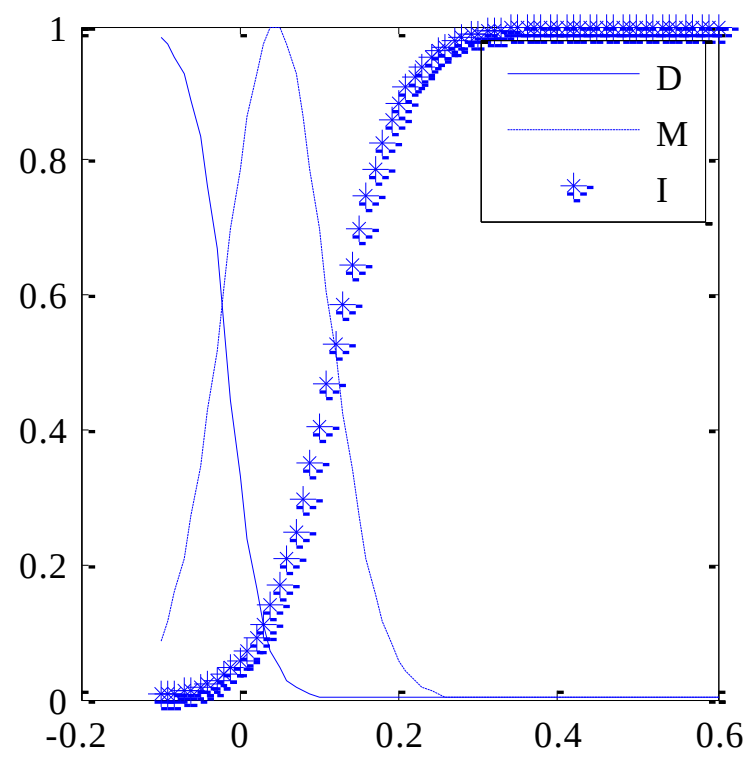

Figure 3: Initial membership functions of $\Delta R_i$

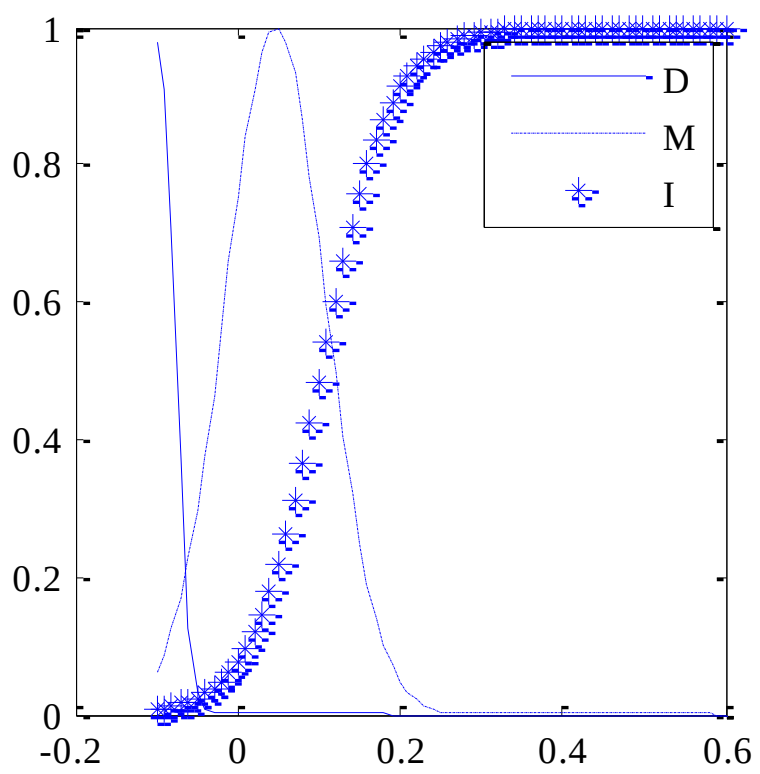

Figure 4: Learned membership functions of $\Delta R_i$

*B. Simulation results*

In the first comparison, the initial deviations of the displacement and the orientation of the robot are chosen to be small: $\sigma_{\Delta S}$ = 2cm, $\sigma_{\Delta\theta}$ = $1^0$. The trajectory of the robot in this simulation is shown in figure 5.

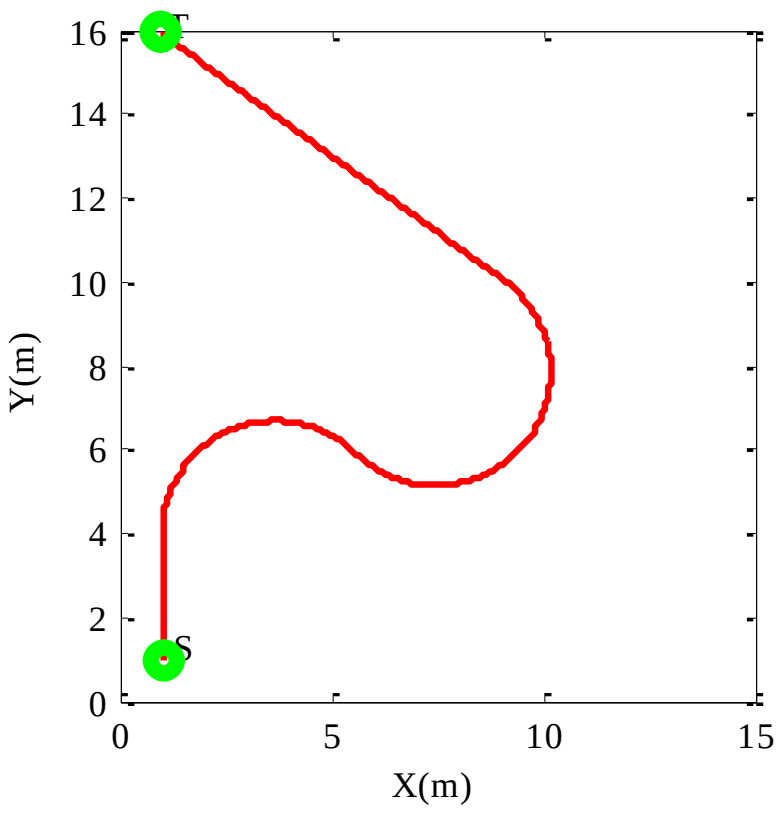

Figure 5: Trajectory of the robot in the first simulation

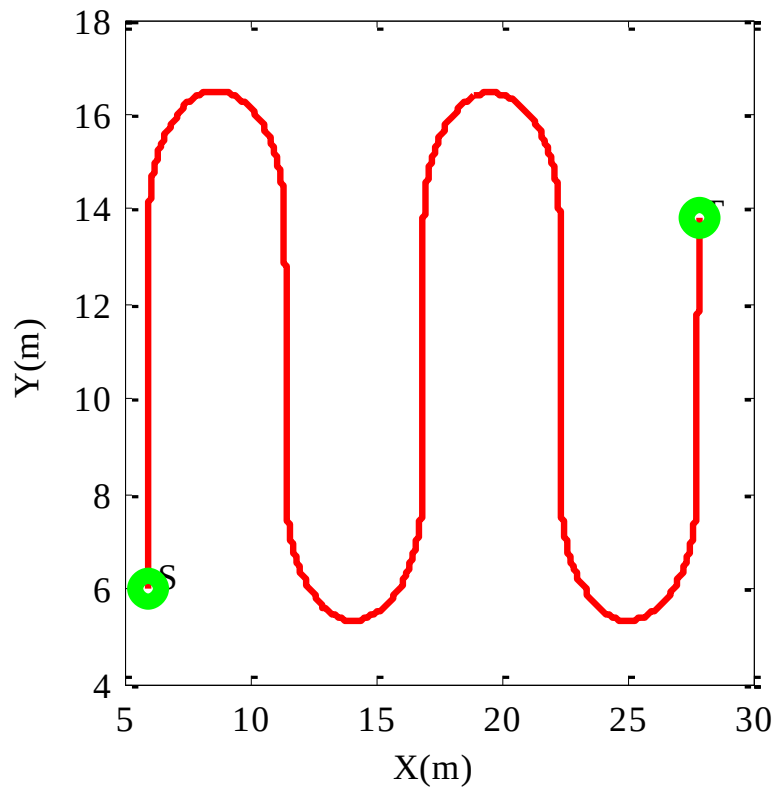

Figure 6: Trajectory of the robot in the second simulation

Figure 7 shows the comparative result of paths estimated by the EKF and FNN-EKF. Noting that only 100 sample points are shown for the convenience of view and comparison. Figure 8-10 show the root mean square error (RMSE) of estimations in the X, Y, and θ direction. It indicates that the FNN-EKF estimation introduces smaller error than the EKF. In the second Monte Carlo simulation, the initial deviations of the displacement and the orientation of the robot are chosen to be l: $\sigma_{\Delta S}$ = 5cm, $\sigma_{\Delta\theta}$ = $3^0$. The trajectory of the robot in this simulation is shown in figure 6. Figure 11-14 show the results. They also indicate that the FNN-EKF estimation is better than than the EKF.

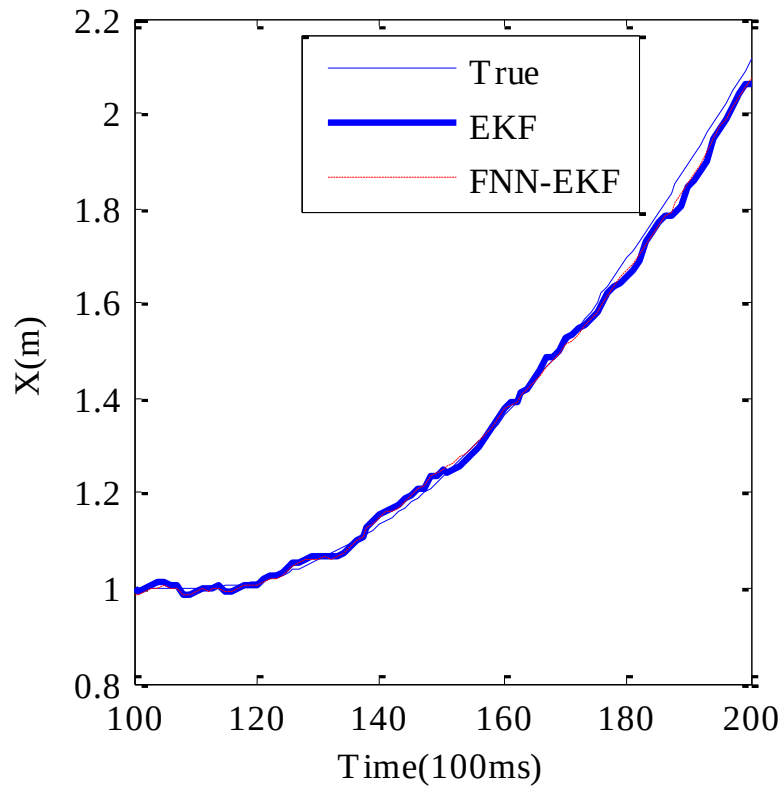

Figure 7: Comparison between the EKF, the FNN-EKF and the true path

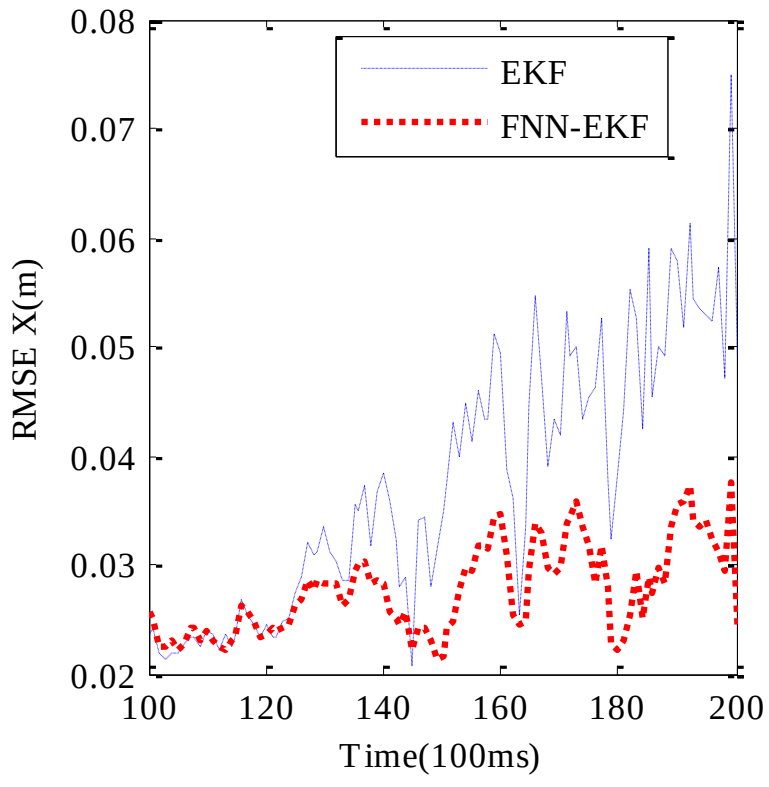

Figure 8: The RMSE of estimations in X direction

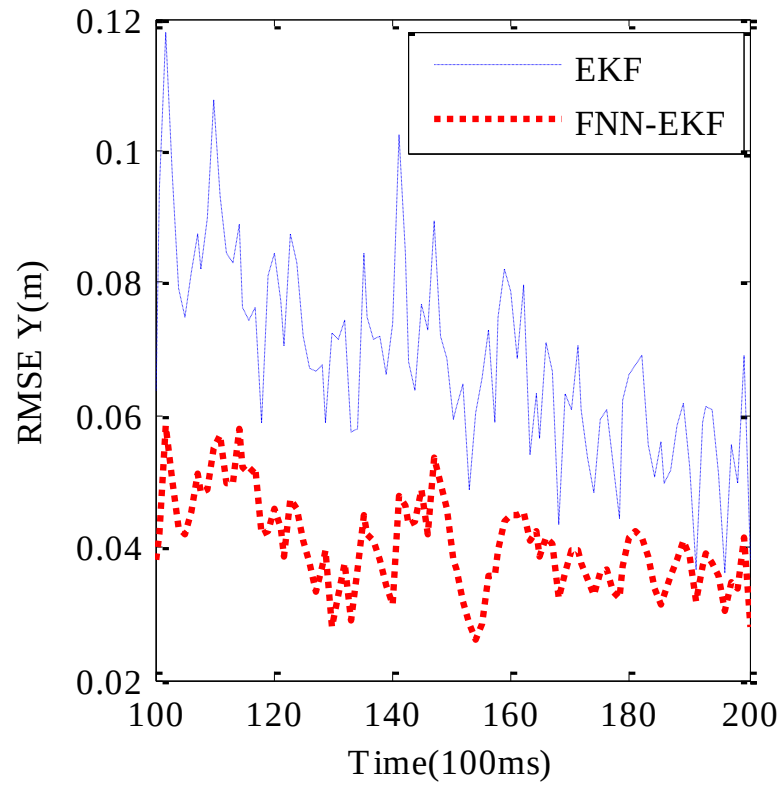

Figure 9: The RMSE of estimations in Y direction

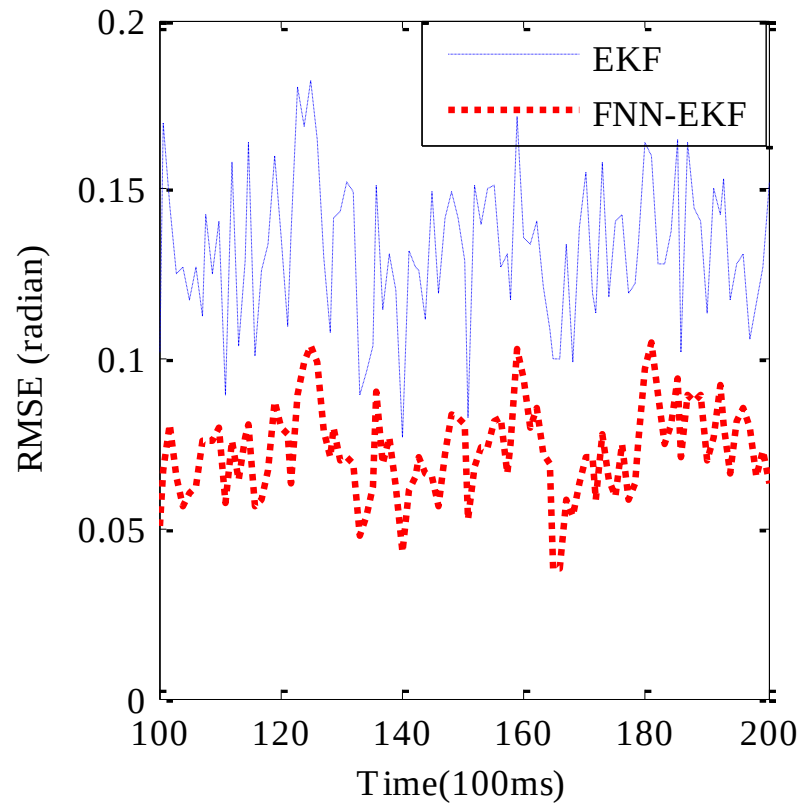

Figure 10: The RMSE of estimations

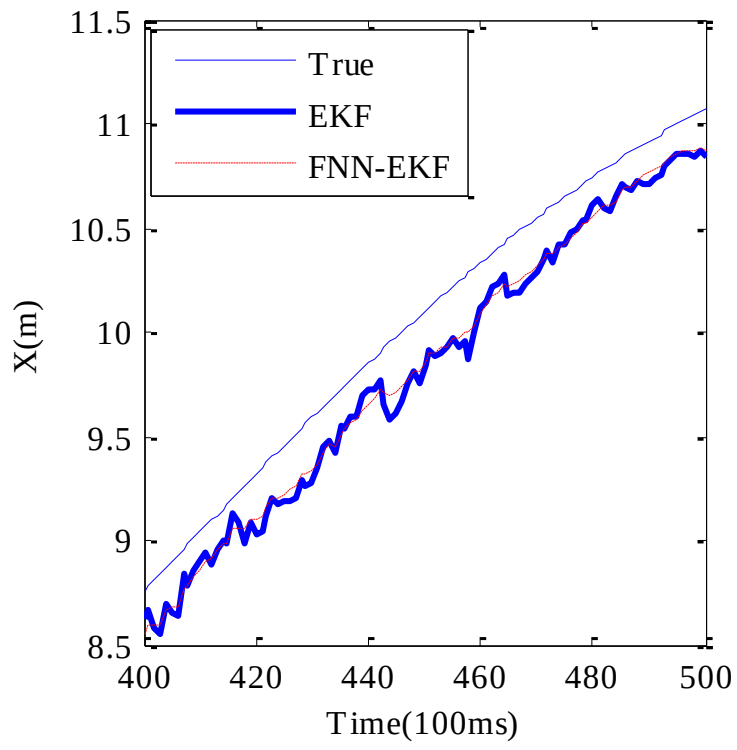

Figure 11: Comparison between the EKF, the FNN-EKF and the true path

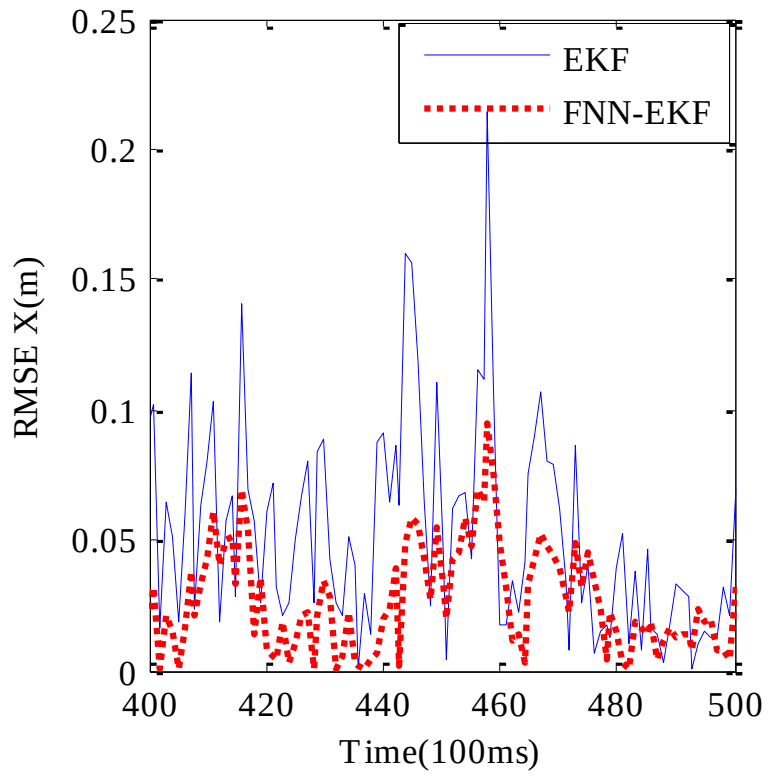

Figure 12: The RMSE of estimations in X direction

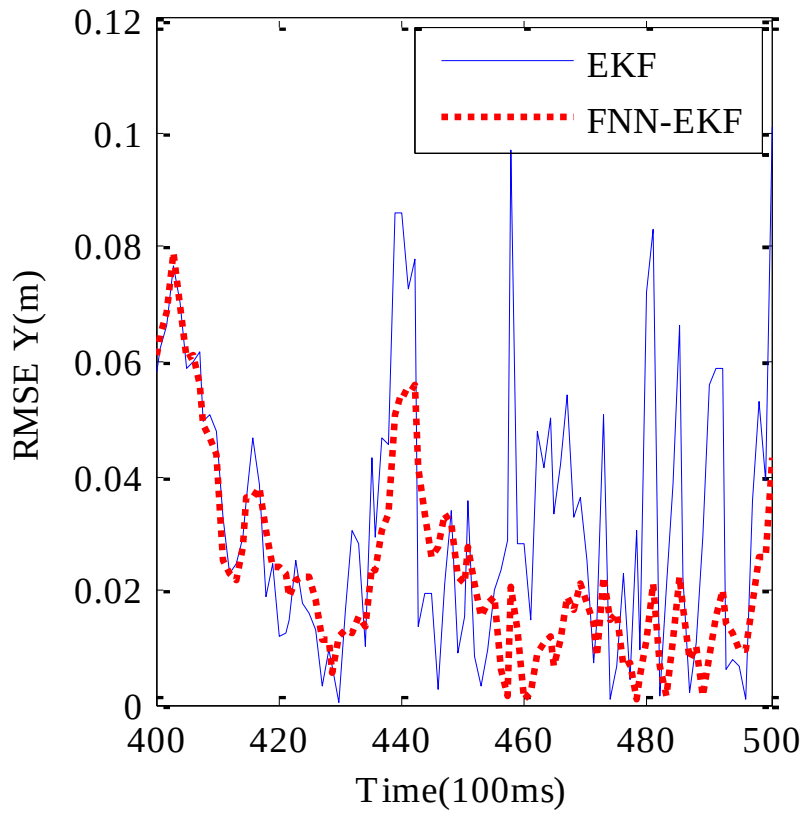

Figure 13: The RMSE of estimations in Y direction

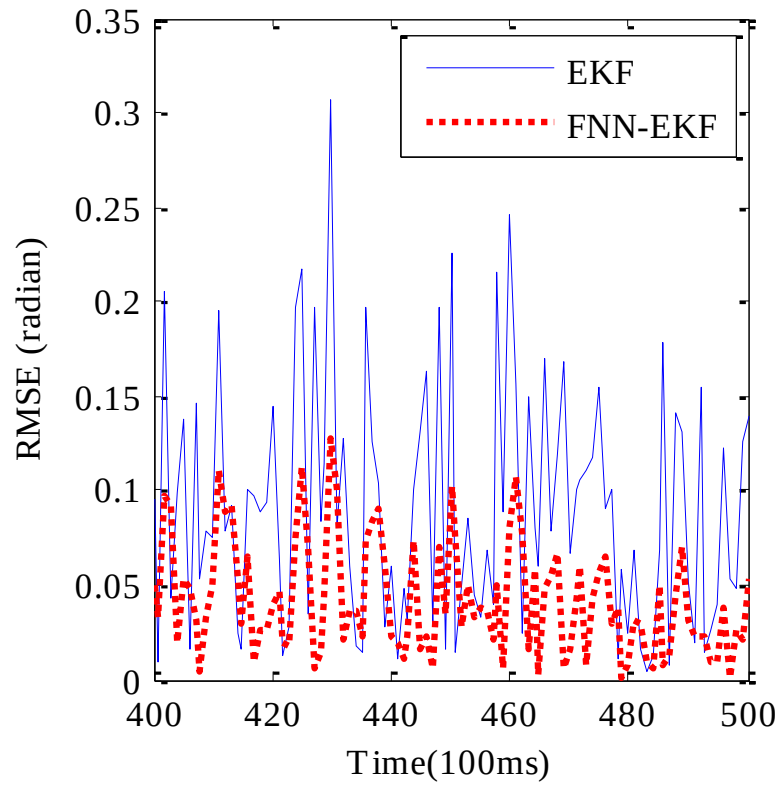

Figure 14: The RMSE of estimations in orientation

## V. EXPERIMENTS

### A. *Experiment setup*

Experiments were conducted in a real mobile robot developed by the author's research group (figure 15). The robot is a two wheeled, differential-drive mobile robot. Its wheel diameter is 0.05m and the distance between two drive wheels is 0.6m. The speed stability of motors controlled by the PID algorithm is ±5%. The sampling time $\Delta t$ is 375ms. The operation environment is flat-floor constructed by cement with several wooden plates surrounded (figure 16). Details of the robot system can be refered from our previous work [15]. The initial value of matries **Q** and **R** is similar to the second simulation case.

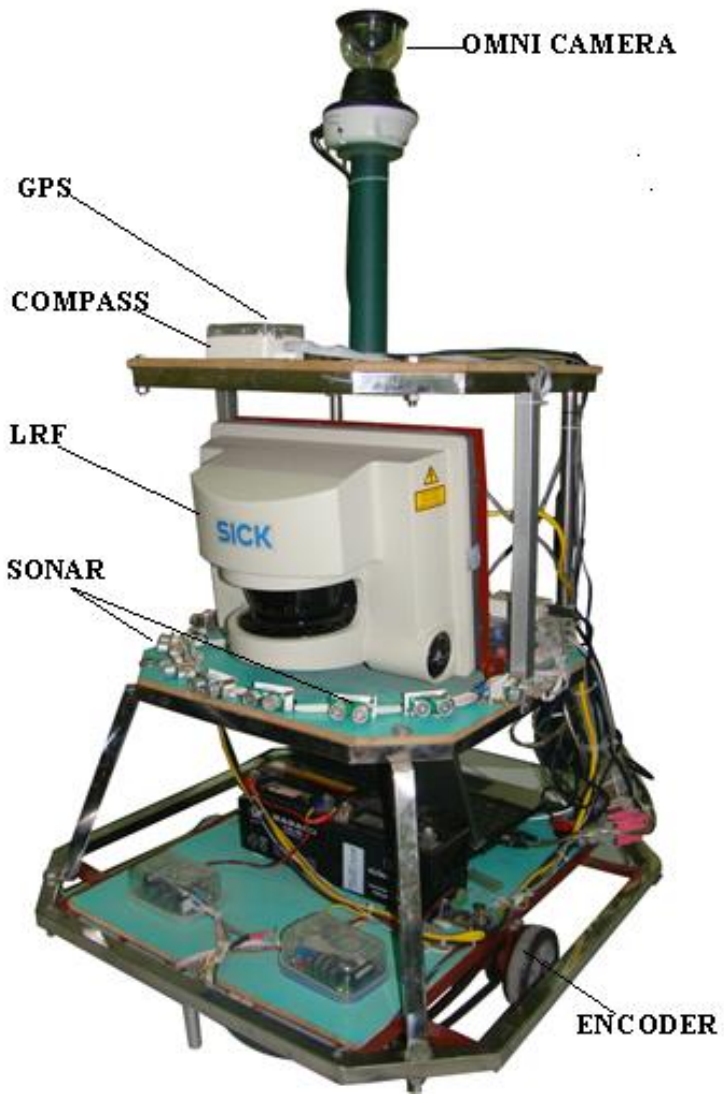

Figure 15: A real mobile robot at laboratory

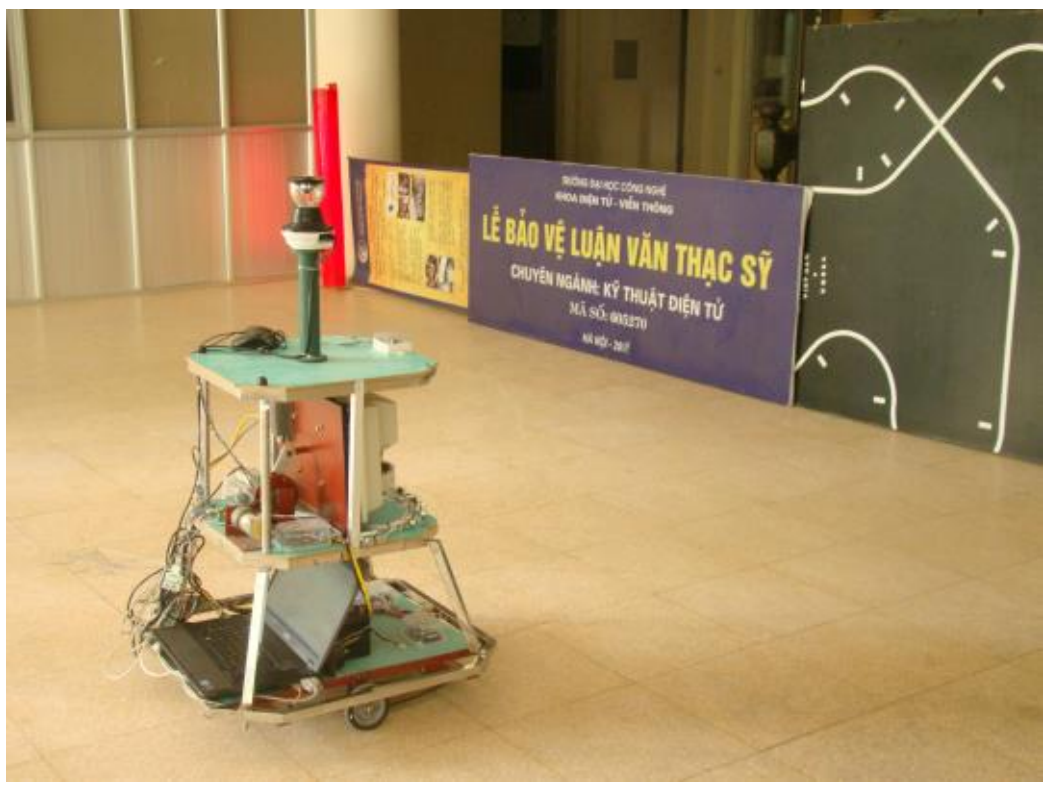

Figure 16. A operation environment of robot

## B. *Experiment results*

In experiments, the robot is controlled to follow a round path. Figure 17 shows the true path and the paths estimated by the EKF and the FNN-EKF. Some parts of trajectory are extracted for the convenience of view as shown in figure 18-20. Due to the asynchronization between experimental values measured on the floor and the values estimated by the filters, it is not possible to compute the RMSE in experiments. Nevertheless, it is able to conclude from figure 18-20 that the FNN-EKF path is closer to the true path than the EKF.

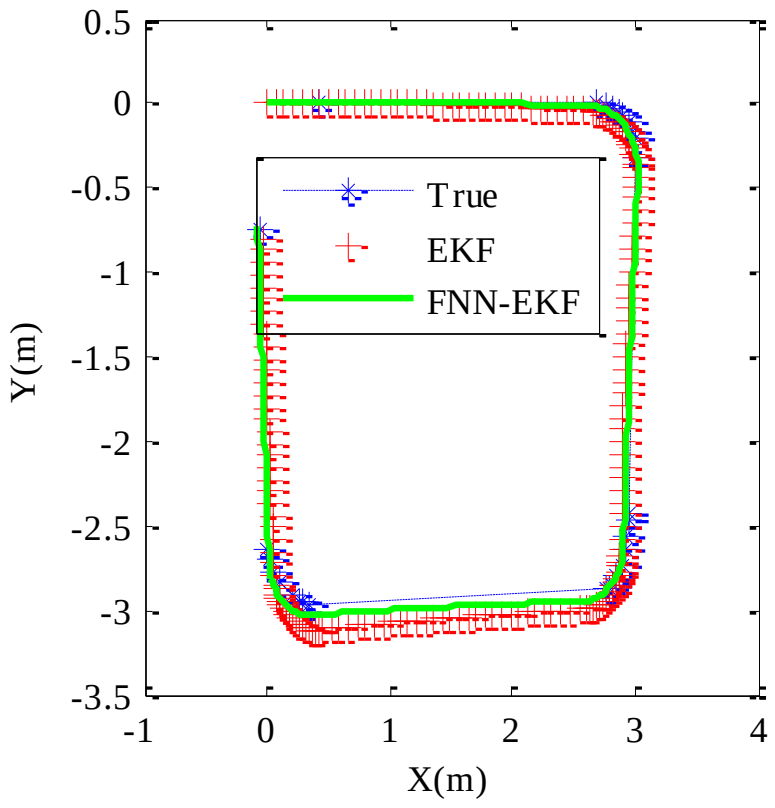

Figure 17: The trajectory of robot

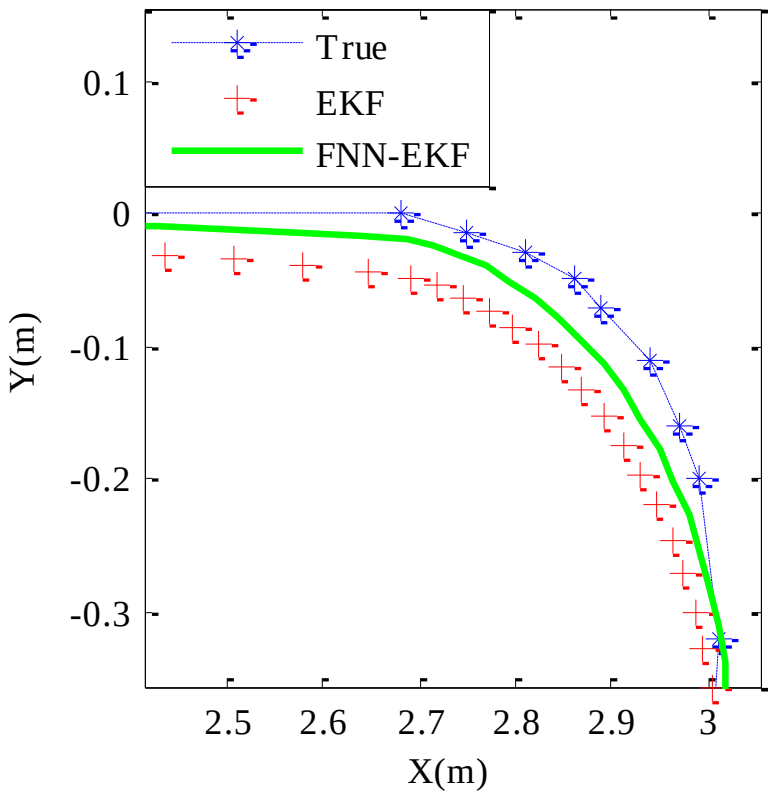

Figure 18: The first corner on the trajectory

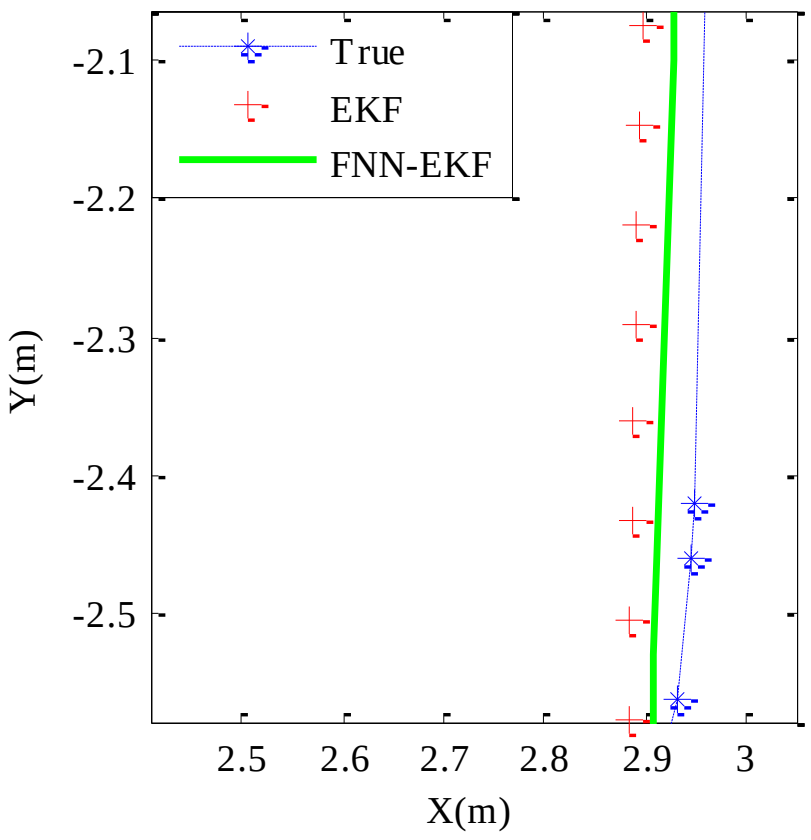

Figure 19: A line on the trajectory

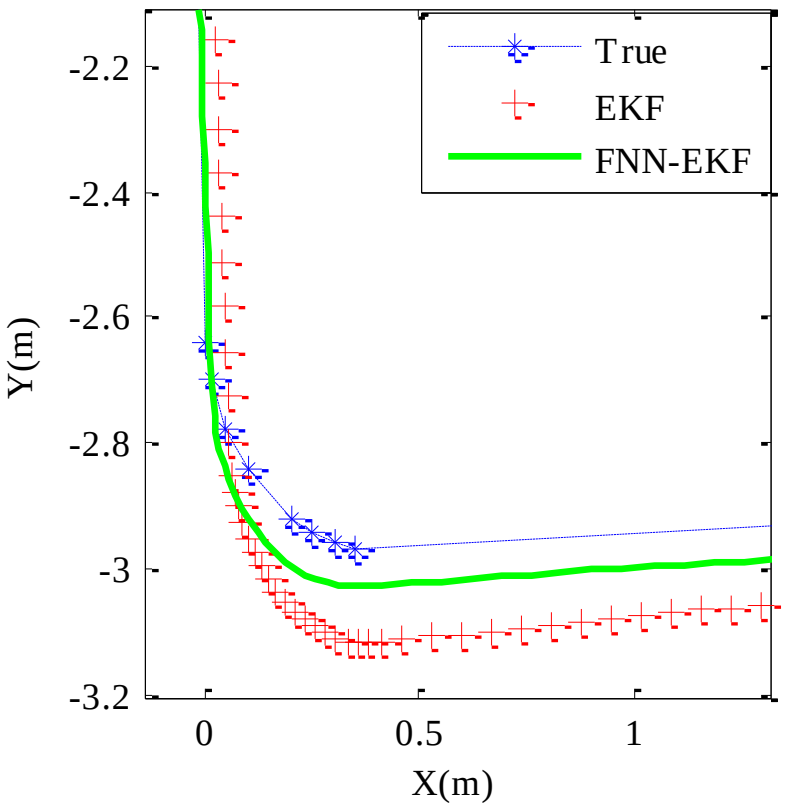

Figure 20: The third corner on the trajectory

## VI. CONCLUSIONS

This paper proposes a novel filter called FNN-EKF for the problem of localization. This filter employs a fuzzy system and a neural network to regulate the noise covariance matrices so that the estimate is converged and more accurate. A number of simulations and experiments have been conducted and the results demonstrate the improvement of the FNN-EKF compared to the EKF. The good localization result in this research can be used as observation data for the controller to control the robot motion in various navigation tasks.

## REFERENCES

[1] Borenstein, J. and Feng, L., 1996. Measurement and correction of systematic odometry errors in mobile robots. *IEEE Transactions on robotics and automation*, *12*(6), pp.869-880.

[2] Borenstein, J., 1994, May. The CLAPPER: A dual-drive mobile robot with internal correction of dead-reckoning errors. In *Proceedings of the 1994 IEEE International Conference on Robotics and Automation* (pp. 3085-3090). IEEE.

[3] Dellaert, F., Fox, D., Burgard, W. and Thrun, S., 1999, May. Monte carlo localization for mobile robots. In *Proceedings 1999 IEEE International Conference on Robotics and Automation (Cat. No. 99CH36288C)* (Vol. 2, pp. 1322-1328). IEEE.

[4] Fox, V., Hightower, J., Liao, L., Schulz, D. and Borriello, G., 2003. Bayesian filtering for location estimation. *IEEE pervasive computing*, *2*(3), pp.24-33.

[5] Grewal, M.S. and Andrews, A.P., 2014. *Kalman filtering: Theory and Practice with MATLAB*. John Wiley & Sons.

[6] Welch, G. and Bishop, G., 1995. An introduction to the Kalman filter.

[7] Duong, P.M., Hoang, T.T., Van, N.T.T., Viet, D.A. and Vinh, T.Q., 2012, July. A novel platform for internet-based mobile robot systems. In *2012 7th IEEE Conference on Industrial Electronics and Applications (ICIEA)* (pp. 1972-1977). IEEE.

[8] Phung, M.D., Van Nguyen, T.T., Tran, T.H. and Tran, Q.V., 2013, July. Localization of networked robot systems subject to random delay and packet loss. In *2013 IEEE/ASME International Conference on Advanced Intelligent Mechatronics* (pp. 1442-1447). IEEE.

[9] Phung, M.D., Nguyen, T.V.T., Quach, C.H. and Tran, Q.V., 2010, December. Development of a tele-guidance system with fuzzy-based secondary controller. In *2010 11th International Conference on Control Automation Robotics & Vision* (pp. 1826-1830). IEEE.

[10] Crowley, J.L., 1989, May. World modeling and position estimation for a mobile robot using ultrasonic ranging. In *ICRA* (Vol. 89, pp. 674-680).

[11] T. Skordata, P. Puget, R.Zigman and N.Ayache, "Building 3-D edgelines tracked in an image sequence", in Pro Intell, Autonomous Systems-2, Amsterdam, 1989, pp 907-919

[12] Kalman, R.E., 1960. A new approach to linear filtering and prediction problems.

[13] Zadeh, L.A., 1988. Fuzzy logic. *Computer*, *21*(4), pp.83-93.

[14] Escamilla-Ambrosio, P.J. and Mort, N., 2001, September. Development of a fuzzy logic-based adaptive Kalman filter. In *2001 European Control Conference (ECC)* (pp. 1768-1773). IEEE.

[15] Hoang, T.T., Duong, P.M., Van, N.T.T., Viet, D.A. and Vinh, T.Q., 2012, December. Development of an EKF-based localization algorithm using compass sensor and LRF. In *2012 12th International Conference on Control Automation Robotics & Vision (ICARCV)* (pp. 341-346). IEEE.